\documentclass[preprint,12pt]{elsarticle}

\usepackage{amsmath,amssymb,amsfonts}
\usepackage{algorithmic}

\newcommand{\Real}{\mathbb{R}}
\newcommand{\Natural}{\mathbb{N}}

\newtheorem{defn}{Definition}
\newtheorem{thm}{Theorem}
\newtheorem{rem}{Remark}
\newtheorem{cor}{Corollary}

\newtheorem{algorithm}{Algorithm}

\newtheorem{assume}{Assumption}

\newcommand{\bfx}{\boldsymbol{x}}

\newcommand{\bfz}{\boldsymbol{z}}
\newcommand{\bfw}{\boldsymbol{w}}

\newcommand{\bfy}{\boldsymbol{y}}

\newcommand{\Cov}{\mathrm{Cov}}

\newcommand \refSec[1]{Section \ref{#1}}

\journal{Information Sciences}

\begin{document}

\begin{frontmatter}



\title{Blessing of dimensionality at the edge}


\author[1,2,3]{Ivan Y. Tyukin\corref{cor1}}
\ead{I.Tyukin@le.ac.uk}
\cortext[cor1]{Corresponding author}

\author[1,2]{Alexander N. Gorban}
\ead{A.N.Gorban@le.ac.uk}

\author[1]{Alistair A. McEwan}
\ead{alistair.mcewan@le.ac.uk}

\author[4]{Sepehr Meshkinfamfard}
\ead{s.meshkinfamfard@ucl.ac.uk}

\author[5]{Lixin Tang}
\ead{lixintang@ise.neu.edu.cn}

\address[1]{University of Leicester, UK}
\address[2]{Lobachevsky University, Russia}
\address[3]{St Petersburg State Electrotechnical University, Russia}
\address[4]{University College London, UK}
\address[5]{ Key Laboratory of Data Analytics and Optimization for Smart Industry (Northeastern University), Ministry of Education, People Republic of China}

\begin{abstract}

In this paper we present theory and algorithms enabling classes of  Artificial Intelligence (AI) systems to continuously and incrementally improve with a-priori quantifiable guarantees -- or more specifically remove classification errors -- over time. This is distinct from state-of-the-art machine learning, AI, and software approaches. Another feature of this approach is that, in the supervised setting, the computational complexity of training is linear in the number of training samples. At the time of classification, the computational complexity is bounded by few inner product calculations. Moreover, the implementation is shown to be very scalable. This makes it viable for deployment in applications where computational power and memory are limited, such as embedded environments.  It enables the possibility for fast on-line optimisation using improved training samples.  The approach is based on the concentration of measure effects and stochastic separation theorems and is illustrated with an example on the identification faulty processes in Computer Numerical Control (CNC) milling and with a case study on adaptive removal of false positives in an industrial video surveillance and analytics system.

\end{abstract}

%

\begin{keyword}
Stochastic separation theorems
\sep artificial intelligence
\sep machine learning
\sep computer vision

\end{keyword}

\end{frontmatter}


\section{Introduction}
\label{sec:introduction}

The past decade has seen extraordinary growth and advances in technologies for collecting and processing very large data streams and data sets.  Central to these advances has been the field of  Artificial Intelligence (AI), built on Machine Learning (ML) and Data Analytics theories.  Exploitation of AI is becoming overwhelmingly ubiquitous.  For instance, end users and consumers use mobile phone apps with AI capabilities, security systems may employ AI to identify unwanted intrusions and infringements, heathcare systems may use AI to assist clinical diagnosis or processes, and mechatronic systems may use AI to implement control including autonomous and semi-autonomous functionality.  Examples of these in literature include those reported in \cite{Liang:2019}, \cite{8574003}.  Whilst this increase in application areas is due in part to advances in AI and ML, it is also due to advances in hardware and supporting platforms.  The emergence of devices such as Nvidia GPUs and Google TPUs have meant that power of server- or super-computer platforms are no longer necessarily required for deployment of deep learning systems.

Whilst state-of-the-art AI systems are capable of vastly outperforming both human and other data mining approaches in identifying minute patterns in very large data sets, their conclusions are vulnerable to data inconsistencies, poor data quality, and the uncertainty inherent to any data.  This uncertainty, together with engineering constraints on implementation and systems integration, leads to inevitable and unavoidable errors.

Consequences of error resulting from AI range from minor inconveniences to safety-critical risks: incorrect cancer treatment options by IBM Watson and several Tesla and Uber crashes in 2018 are a few examples of the latter.  However the solution to ameliorating or eliminating errors is non-trivial.  The field of Software Engineering has provided numerous approaches for understanding the behaviours and misbehaviours of software based systems ranging from efficient scenario based testing techniques through to formal verification---see for instance \cite{8369576}. However these software architectures typically do not contain the inherent uncertainties of data driven AI---although this is rapidly changing with the push towards higher levels of driver assist and autonomy.  Recent examples that consider autonomous vehicle control systems that incorporating AI include \cite{7993050} and \cite{8723533}, although common to these approaches is to look at systems level behaviours rather than the correctness of the AI component.

Whilst structuring data, improving the quality of data, and removing uncertainty is known to improve quality, it is too resource intensive in the general case and thus unsustainable across sectors and industries.  Moreover, whilst it may improve the quality of high-assurance or safety-critical systems, it does not provide a measure of understanding of quality and consistency of output.  More fundamentally, constraints on implementations such as quantization errors and memory limits present challenges to AI performance in resource constrained embedded settings---referred to as ``at the edge''.

\subsection{Background and related work}
\label{sec:background}

Significant efforts have been applied to address errors in AI systems. Using ensembles \cite{Hansen:1990}, \cite{Ho:1995}, \cite{Ho:1998}, augmenting training data  \cite{Kuznetsova:2015}, \cite{Misra:2015}, \cite{Prest:2012}, enforcing continuity \cite{Zheng:2016}, and AI knowledge transfer \cite{Pratt:ANIP:1992}, \cite{chen2015net2net}, \cite{vapnik2017knowledge} have been extensively discussed in the literature. These measures, however, do not warrant error-free behaviour as AIs based on empirical data are expected to make mistakes.

Recently in \cite{GorTyuRom2016}, \cite{GorbanTyukin:NN:2017}, \cite{tyukin2017knowledge}, \cite{gorban2018correction}, \cite{GorTyuRom2019}, \cite{GorTyuGreen2019} we have shown that spurious errors of AI systems operating in high-dimensional spaces (convolutional and deep learning neural networks being the canonical examples) can be efficiently removed by Fisher discriminants. The advantage of this approach over, for instance, Support Vector Machines (SVM) \cite{Vapnik2000} is that the computational complexity for constructing Fisher discriminants is at most linear in the number of points in the training set whereas the worst-case complexity for SVM scales as a cubic function of the training set size \cite{chapelle2007training}.

This method is applicable to identified singular spurious errors as well as to moderate-sized clusters.   The question then naturally arises as to what happens if the volume of errors produced is similar to the volume of correct responses?  Moreover, if it is possible for a deployed AI to keep improving its performance with limited resources available for supervision and re-training?   Both of these questions are fundamentally relevant across the spectrum of AI applications. Notably, given the computational complexity properties that they enjoy,  they are particularly acute for embedded and resource-constrained systems often referred to as ``at the edge''.

\subsection{Contribution and structure of this paper}

In this paper we show that stochastic separation theorems, or the blessing of dimensionality \cite{gorban2018blessing}, \cite{kainen1997utilizing},  stemming from the concentration of measure effects \cite{GorTyu:2016}, \cite{Kurkova}, \cite{GAFA:Gromov:2003}, can be adapted and applied to address these questions. We present and justify both mathematically and experimentally an algorithm capable of delivering the removal of errors at computational costs compatible with deployment at the edge. The algorithm has both supervised and unsupervised components which enables it to adapt to data without additional supervisory inputs. As compared to previously proposed stochastic separation-based algorithms \cite{tyukin2017knowledge}, \cite{GorTyuGreen2019}, current algorithm was shown to consistently learn from large volumes of errors making its deployment in applications with uncertainty and bias in training data particularly attractive.


The paper is organized as follows. \refSec{sec:notation} sets out the notation we use in this paper.  \refSec{sec:problem} contains necessary theoretical preliminaries and formal statement of the problem. In \refSec{sec:theory} we present a new algorithm for improving AIs ``at the edge''.  \refSec{sec:examples} illustrates an application of the proposed algorithms in two industrial applications: product quality prediction in milling machines and automated edge-based object detection in large-scale survaliance systems; \refSec{sec:conclusion} concludes the paper.

\section{Notation}
\label{sec:notation}

The following notational agreements are used throughout the text:

\begin{itemize}
	\item $\Real^n$ stands for the $n$-dimensional linear real vector space, and $\Real_{\geq 0}=\{x\in \Real | \ x\geq 0\}$;
	\item $\Natural$ denotes the set of natural numbers;
	\item symbols $\boldsymbol{x} =(x_{1},\dots,x_{n})$ will denote elements of $\Real^n$;
	\item $(\boldsymbol{x},\boldsymbol{y})=\sum_{k} x_{k} y_{k}$ is the inner product of $\boldsymbol{x}$ and $\boldsymbol{y}$, and $\|\boldsymbol{x}\|=\sqrt{(\boldsymbol{x},\boldsymbol{x})}$ is the standard Euclidean norm  in $\Real^n$;	\item  $\mathbb{B}_n(r,\bfy)$  stands for the ball in $\Real^n$ of radius $r$ centered at $\bfy$ : $\mathbb{B}_n(r,\bfy)=\{\boldsymbol{x}\in\Real^n| \ \left(\boldsymbol{x}-\bfy,\boldsymbol{x}-\bfy\right)\leq r^2\}$;
	\item  $\mathbb{B}_n$ denotes for the unit ball in $\Real^n$ centered at the origin: $\mathbb{B}_n=\{\boldsymbol{x}\in\Real^n| \ \left(\boldsymbol{x},\boldsymbol{x}\right)\leq 1\}$;
	\item $V_n$ is the $n$-dimensional Lebesgue measure, and $V_n(\mathbb{B}_n)$ is the volume of unit ball;
	\item if $\mathcal{Y}$ is a finite set then the number of elements in $\mathcal{Y}$ (cardinality of $\mathcal{Y}$) is denoted by $|\mathcal{Y}|$.
\end{itemize}

\section{Problem Formulation and Mathematical Preliminaries}
\label{sec:problem}

\subsection{Problem formulation}

Following \cite{GorTyuRom2019}, we consider  a generic AI system that processes some {\it input} signals, produces {\it internal} representations of the input and returns some {\it outputs}. We assume that there is a sampling process whereby some relevant information about the input, internal signals, and outputs are combined into a common vector, $\bfx$, representing, but not necessarily defining, the {\it state} of the AI system.

Depending on the sampling process, the vector $\bfx$ may have various numbers of elements. But generally, the objects $\bfx$ are assumed to be elements of $\Real^n$, with $n$ depending on the sampling process. Over a period of activity the AI system generates a set $\mathcal{X}=\{\bfx_1,\dots,\bfx_M\}$ of representations $\bfx$. In agreement with  standard assumptions in machine learning literature \cite{Vapnik2000}, we assume that the set $\mathcal{X}$ is is a random sample drawn from some {distribution}. The distribution that generates vectors $\bfx$ is supposed to be {\it unknown}.   We will, however, impose some mild technical assumption on the generating probability distribution.

The central question we would like to address here is how to create algorithms capable of producing a single or an esemple  of linear functionals suitable for decision-making at-the-edge. The focus on linear functionals is motivated by computational efficiency of their implementation in embedded settings. We would like to avoid using the framework of Support Vector Machines due to the computational costs involved which may present an obstable for embedded deployment.

\begin{assume}\label{assume:bounded_pdf} The probability density function, $p$, associated with the probability distribution of the random variable $\bfx$ exists, is defined on the unit ball $\mathbb{B}_n$, and there exist $C>0$ and $r\in (0,2)$ such that
\begin{equation}\label{eq:non-concentration:0}
\int_{\mathbb{B}_n(1/2,\bfz)} p(\bfx)d\bfx \leq C\left(\frac{r}{2}\right)^n \ \mbox{for all} \ \bfz\in\mathbb{B}_n, \ \|\bfz\|=1/2.
\end{equation}
\end{assume}

The assumption requires that the random variable $\bfx$ is in $\mathbb{B}_n$ which is consistent with the scope of our applications. The other part of the assumption, condition (\ref{eq:non-concentration:0}), is a version of the Smeared Absolute Continuity (SmAC) property introduced in \cite{gorban2018correction}, \cite{gorban2018augmented}.  Awareness of the latter property will be important for the algorithms that follow.  In addition to Assumption \ref{eq:non-concentration:0} it will be convenient to consider alternative specifications of the data probability distribution which are captured in Assumption \ref{assume:non-concentration:1} below.

\begin{assume}\label{assume:non-concentration:1}  The probability density function, $p$, associated with the probability distribution of the random variable $\bfx$ exists, is defined on the unit ball $\mathbb{B}_n$, and there exist $C>0$ and $r\in\Real_{\geq 0}$ such that
\begin{equation}\label{eq:non-concentration}
 p(\bfx) < \frac{C r^n}{V_n(\mathbb{B}_n)}, 
\end{equation}
for all $\bfx\in\mathbb{B}_n$.
\end{assume}

Note that if  Assumption \ref{assume:non-concentration:1} holds with $r\in(0,2)$ then it automatically implies that Assumption \ref{assume:bounded_pdf} holds true too. In case of Assumption \ref{assume:non-concentration:1} we do not wish yet to specify exact values of $C$ and $r$. We would, however, like to formalise a particularly useful form of its dependence on dimension $n$ and the volume of the unit ball $\mathbb{B}_n$ captured by (\ref{eq:non-concentration}).

\begin{defn}\label{Def:OneSep} {\em A point $\boldsymbol{x}\in  \mathbb{R}^n$ is linearly separable}  from a set $\mathcal{Y} \subset \mathbb{R}^n$, if there exists a linear functional $l(\cdot)$ such that
	\[
	l(\bfx) > l(\bfy)
	\]
	for all $\boldsymbol{y}\in \mathcal{Y}$.
\end{defn}

\begin{defn}\label{Degf:SepOfSets} {\em A set $\mathcal{X} \subset  \mathbb{R}^n$ is linearly separable from a set $\mathcal{Y} \subset \mathbb{R}^n$},  if there exists a linear functional $l(\cdot)$ such that
	\[
	l(\bfx) > l(\bfy)
	\]
	for all $\boldsymbol{y}\in \mathcal{Y}$ and $\bfx\in\mathcal{X}$.
\end{defn}

In addition to these standard notions of linear separability, we adopt the notion of Fisher separability \cite{gorban2018correction}, \cite{gorban2018augmented}. Observe that the classical Fisher separability requires Maholonoibs inner product or whitening. 

\begin{defn}\label{Def:FishSep} {\em A point $\boldsymbol{x}\in  \mathbb{R}^n$ is Fisher separable}  from a set $\mathcal{Y} \subset \mathbb{R}^n$, if
	\begin{equation}\label{eq:FishSep}
	(\bfx,\bfx) > (\bfx,\bfy)
	\end{equation}
	for all $\boldsymbol{y}\in \mathcal{Y}$. The point is Fisher separable from the set $\mathcal{Y}$ with a threshold $\kappa\in[0,1)$ if
	\begin{equation}\label{eq:FishSep:threshold}
	(\bfx,\bfx) > \kappa(\bfx,\bfy)
	\end{equation}
\end{defn}

Having introduced all relevant assumptions and notions, we are now ready to proceed with results underpinning our algorithmic developments.

\subsection{Mathematical Preliminaries}

Our first result is provided in Theorem \ref{thm:one_separability} (cf. \cite{GorMakTyu:2018}, \cite{gorban2018correction})

\begin{thm}\label{thm:one_separability} Let $\mathcal{X}=\{\bfx_1,\dots,\bfx_M\}$  be given, $\bfx_i\in\mathbb{B}_n$, and let $\bfx$ be drawn from a distribution satisfying Assumption \ref{assume:bounded_pdf}. Then $\bfx$ is Fisher separable from  the set $\mathcal{X}$ with probability
	\begin{equation}\label{eq:FishSep:probability:1}
	P \geq 1 - M C \left(\frac{r}{2}\right)^n, \ r\in(0,2).
	\end{equation}
\end{thm}
{\it Proof of Theorem \ref{thm:one_separability}}. Consider events
\[
A_i: \ \bfx \ \mbox{is Fisher separable from} \ \bfx_i.
\]
According to Definition \ref{Def:FishSep}, this is equivalent to that $(\bfx,\bfx)-(\bfx,\bfx_i)>0$. Therefore
\[
P(\mbox{not} \ A_i) = \int_{(\bfx,\bfx)-(\bfx,\bfx_i)\leq 0} p(\bfx)d\bfx.
\]
According to the De Morgan's law,
\[
{\bigwedge_{i=1}^M A_i} = \mbox{not} \ \left({\bigvee_{i=1}^M (\mbox{not} \ {A_i}) }\right).
\]
Hence
\[
P(A_1 \land  A_2 \land \cdots \land A_M)=1-P( (\mbox{not} \  A_1)\lor (\mbox{not} \ A_2) \lor \cdots (\mbox{not}  \ A_M)),
\]
and consequently
\begin{equation}\label{eq:prob_bound}
P(A_1 \land A_2 \land \cdots \land A_M)\geq 1 - \sum_{i=1}^M P(\mbox{not} \ A_i).
\end{equation}

Consider the set
\[
\Omega_i=\{\bfx\in\mathbb{B}_n \ | (\bfx,\bfx)-(\bfx,\bfx_i)\leq 0 \} = \mathbb{B}_n(\|\bfx_i\|/2,\bfx_i/2).
\]
Since  $\bfx_i\in\mathbb{B}_n$, it follows that  $\Omega_i\subseteq \mathbb{B}_n(1/2,\bfx_i/2)$. This and Assumption \ref{assume:bounded_pdf} imply that
\begin{equation}\label{eq:proof:prob_ball_est:1}
\int_{(\bfx,\bfx)-(\bfx,\bfx_i)\leq 0} p(\bfx)d\bfx \leq \int_{\mathbb{B}_n(1/2,\bfx_i/2)} p(\bfx)d\bfx \leq C \left(\frac{r}{2}\right)^n
\end{equation}
Combining (\ref{eq:proof:prob_ball_est:1}) and  (\ref{eq:prob_bound}) we can conclude that the probability that $\bfx$ is separable from all $\bfx_i$ is bounded from below by the expression in (\ref{eq:FishSep:probability:1}).  $\square$

\begin{rem}\normalfont According to Theorem \ref{thm:one_separability}, a single-point set $\mathcal{Y}$ under some mild hypotheses can be separated from $\mathcal{X}$ by a simple Fisher discriminant with probability close to one. Denoting 
\[
\delta = M C \left(\frac{r}{2}\right)^n
\]
one can conclude that $\mathcal{Y}$ is Fisher separable from a given set $\mathcal{X}\subset \mathbb{B}_n$ with probability grater or equal to $1-\delta$ if
\[
n\geq \frac{ \log M + \log C - \log{\delta}}{\log 2 - \log r}.
\]
For $M=10^4$, $C=10^2$, $\delta=10^{-3}$, and $r=1$, the above inequality holds for all $n\geq 30$.  
\end{rem}

In practice, however, we may be interested in a situation when the set $\mathcal{Y}$ contains several elements which may or may not have some inherent clustering structure or concentrations.  In what follows we derive a generalization of Theorem \ref{thm:one_separability} enabling us to formally address the latter case too. 

Consider two random sets $\mathcal{X}=\{\bfx_1,\dots,\bfx_M\}$ and $\mathcal{Y}=\{\bfy_1,\dots,\bfy_{K}\}$. Let there be a process (e.g. a learning algorithm) which, for the given $\mathcal{X}$, $\mathcal{Y}$ or their subsets, produces a classifier
\[
f(\cdot)=\sum_{i=1}^d \alpha_i  (\bfz_i,\cdot), \ \alpha_j\in\Real.
\]
The vectors $\bfz_i$, $i=1,\dots,d$ are supposed to be known. Furthermore, we suppose that the function $f$ is such that
\begin{equation}\label{eq:assume:alpha:0}
f(\bfy_j)>\sum_{m,k=1}^d \alpha_m \alpha_k (\bfz_m,\bfz_k)
\end{equation}
for all $\bfy_j\in\mathcal{Y}$. In other words, if we denote $\bfw=\sum_{i=1}^d \alpha_i \bfz_i$, the following holds true:
\begin{equation}\label{eq:assume:alpha:1}
(\bfw,\bfw)<(\bfw, \bfy_i) \ \mbox{for all} \ i=1,\dots, K.
\end{equation}
Note that since the $\mathcal{Y},\mathcal{X}$ are random, it is natural to expect that the vector $\boldsymbol{\alpha}=(\alpha_1,\dots,\alpha_d)$ is also random.  The following statement can now be formulated:

\begin{thm}\label{thm:set_separability} Consider sets $\mathcal{X}=\{\bfx_1,\dots,\bfx_M\}$ and $\mathcal{Y}=\{\bfy_1,\dots,\bfy_{K}\}$. Let $p_\alpha(\boldsymbol{\alpha})$ be the probability density function associated with the random vector $\boldsymbol{\alpha}$, and $\boldsymbol{\alpha}$ satisfies condition (\ref{eq:assume:alpha:0}) with probability $1$. Then the set $\mathcal{X}$ is separable from  the set $\mathcal{Y}$ with probability
	\begin{equation}\label{eq:SetSep:probability:1}
	P\geq 1 - \sum_{i=1}^M \int_{H(\boldsymbol{\alpha},\bfx_i)\leq 0} p_\alpha(\boldsymbol{\alpha})d\boldsymbol{\alpha},
	\end{equation}
	where
	\[
	H(\boldsymbol{\alpha},\bfx_i)=\sum_{k,m=1}^d \alpha_k\alpha_m (\bfz_k,\bfz_m)-\sum_{m=1}^d \alpha_m  (\bfz_m,\bfx_i).
	\]
\end{thm}
{\it Proof of Theorem \ref{thm:set_separability}}.  Consider events
\[
A_i: \ (\bfw,\bfw)>(\bfw,\bfx_i).
\]
Event $A_i$ is equivalent to that the inequality
\[
H(\boldsymbol{\alpha},\bfx_i)=\left(\sum_{k=1}^d \alpha_k \bfz_k, \sum_{k=m}^d \alpha_m \bfz_m\right)-\sum_{m=1}^d \alpha_m (\bfz_m, \bfx_i)>0
\]
holds true. According to (\ref{eq:prob_bound}),  equation (\ref{eq:SetSep:probability:1}) is a lower bound for the probability that all these events hold. Recall that vectors $\boldsymbol{\alpha}$ satisfy (\ref{eq:assume:alpha:1}), and hence
\[
\begin{aligned}
\sum_{m=1}^d \alpha_m (\bfz_m,\bfx_i)=&(\bfw,\bfx_i)\\
&<(\bfw,\bfy_j)=\sum_{m=1}^d \alpha_m (\bfz_m,\bfy_j)
\end{aligned}
\]
for all $\bfx_i\in\mathcal{X}$ and $\bfy_j\in\mathcal{Y}$ with probability at least (\ref{eq:SetSep:probability:1}). The statement now follows immediately from Definition \ref{Degf:SepOfSets}. $\square$

\begin{rem} \normalfont Theorem \ref{thm:set_separability} generalizes earlier $k$-tuple separation theorems \cite{tyukin2017knowledge} to a very general class of practically relevant distributions. No independence assumptions are imposed on the components of vectors $\bfx_i$  and $\bfy_i$. We do, however, require that some information about distribution of the classifier parameters, $\boldsymbol{\alpha}$, is available.

Observe, for example, that if there exist $L>0$, $\lambda\in (0,1)$ and a function $\beta:\Natural\times \Natural \rightarrow \Real$ such that
\[
\int_{H(\boldsymbol{\alpha},\bfy)\leq 0} p_\alpha(\boldsymbol{\alpha})d\boldsymbol{\alpha} \leq L \lambda^{\beta(d,n)}
\]
for any $\bfy\in\Real^n$ then (\ref{eq:SetSep:probability:1}) becomes
\[
P\geq 1- M  L \lambda^{\beta(d,n)}.
\]
If $d=n$ and elements of the set $\mathcal{Y}$ are sufficiently strongly correlated, then the above bound becomes similar to (\ref{eq:FishSep:probability:1})  from Theorem \ref{thm:one_separability}. The latter provides a good approximation of the separability probability bound for a simple separating function in which $\bfw$ is just a scaled centroid of the set $\mathcal{Y}$.
\end{rem}

Theorems \ref{thm:one_separability} and \ref{thm:set_separability} link dimensionality of the decision-making space with opportunities for quick and fast learning by mere Fisher discriminants. Before, however, moving on to the actual algorithms for either improving legacy data-driven AI systems or generating new edge-based classifiers or  both, let us examine another useful property of data in high dimension. This property is summarised in Theorem \ref{thm:dichotomy}.

\begin{thm}\label{thm:dichotomy} Let $\mathcal{X}=\{\bfx_1,\dots,\bfx_M\}$ be a finite sample of elements $\bfx_i$ drawn identically and independently from a distribution satisfying Assumption \ref{assume:non-concentration:1}. Let $\bfy$ be an element of $\mathbb{B}_n$. 

Then
\begin{itemize}
\item[1) ] $\bfy$ is Fisher separable from the set $\mathcal{X}$ with probability 
\[
P\geq 1 - M C\left((1-\|\bfy\|^2)^{\frac{1}{2}} r \right)^n
\]

\item[2) ] every $\bfx\in\mathcal{X}$ is Fisher separable from $\bfy$ with probability
\[
P\geq 1 - M C \left(\|\bfy\| r \right)^n.
\]
\end{itemize}
\end{thm}
{\it Proof of Theorem \ref{thm:dichotomy}}. Consider first statement 1) of the theorem. According to Definition \ref{Def:FishSep}, the point $\bfy$ is Fisher separable from $\mathcal{X}$ if 
\[
\|y\|^2=(\bfy,\bfy) > (\bfy,\bfx) \ \forall \ \bfx\in\mathcal{X}
\]
Given that $\mathcal{X}\in\mathbb{B}_n$, this is equivalent to the fact that no element of $\mathcal{X}$ belongs to the spherical cap
\[
\mathcal{C}_n(\bfy)=\left\{\bfz\in\mathbb{B}_n \ | \ \left(\frac{\bfy}{\|\bfy\|},\bfz\right) - \|\bfy\| \geq 0  \right\}.
\]
The probability that an $\bfx\in\mathcal{X}$ ends up in the cap is
\[
\int_{\mathcal{C}_n(\bfy)} p(\bfx) d\bfx \leq  \int_{\mathbb{B}_n(\sqrt{1-\|\bfy\|^2},\bfy)} p(\bfx) d\bfx\leq \left((1-\|\bfy\|^2)^{\frac{1}{2}} r \right)^n.
\]
Combining this with (\ref{eq:prob_bound}) gives the required bound.

Statement 2) follows from the observation that all $\bfx\in\mathcal{X}$ that are outside of the ball $\mathbb{B}_n(\|\bfy\|,0)$ are Fisher separable from $\bfy$:
\[
\|\bfx\| > \|\bfy\| \Rightarrow  (\bfx,\bfx)  - (\bfx,\bfy)  \geq  \|\bfx\| \|\bfy\|  - (\bfx,\bfy) > 0.
\]
$\square$

Theorem \ref{thm:dichotomy} enables us to formulate a simple corollary revealing an interesting dichotomy of datasets in high-dimensional spaces. More precisely

\begin{cor}\label{cor:dichotomy} Let $\mathcal{X}=\{\bfx_1,\dots,\bfx_M\}$ be a finite sample of elements $\bfx_i$ drawn identically and independently from a distribution satisfying Assumption \ref{assume:non-concentration:1} with $r<\sqrt{2}$. Let $\bfy$ be an element of $\mathbb{B}_n$. Then, with probability 
\[
P\geq  1 - M C \left(\frac{r}{\sqrt{2}} \right)^n,
\]
either 
\begin{itemize}
\item[1) ]  $\bfy$ is Fisher separable from $\mathcal{X}$
\end{itemize}
 or 
 \begin{itemize}
 \item[2) ] every $\bfx\in\mathcal{X}$ is Fisher separable from $\bfy$  and  $\bfy$  is inside the ball $\mathbb{B}_n(1/\sqrt{2},0)$. 
 \end{itemize}
 \end{cor}
{\it Proof of Corollary \ref{cor:dichotomy}}. Let $\bfy$ be an arbitrary element of $\mathbb{B}_n$. Then either $1-\|\bfy\|^2 \leq \|\bfy\|^2$ or $1-\|\bfy\|^2 > \|\bfy\|^2$. If the first alternative holds true then $1-\|\bfy\|^2 \leq 1/2$ and statement 1) follows from Theorem \ref{thm:dichotomy}, alternative 1. If the opposite holds true then statement 2) follows from  Theorem \ref{thm:dichotomy}, alternative 2. $\square$ 

Corollary \ref{cor:dichotomy} formalizes an interesting dichotomy of high-dimensional data: {\it for  sufficiently large dimension $n$, a finite sample $\mathcal{X}$ drawn from a class of distributions  (Assumption \ref{assume:non-concentration:1} with $r<\sqrt{2}$ and suitable $C$), with probability close to $1$, a given point in $\mathbb{B}_n$ (regardless of a distribution it has been drawn from) is either inside the ball $\mathbb{B}_n(1/\sqrt{2},0)$  or it is Fisher-separable from the sample $\mathcal{X}$.}  

We will exploit this and other relevant properties formulated in Theorems \ref{thm:one_separability} -  \ref{thm:dichotomy} in the next section.

\section{Fast removal of AI errors}
\label{sec:theory}

Consider two finite sets, the set $\mathcal{X}\subset\Real^n$, and $\mathcal{Y}\subset\Real^n$. The task is to efficiently construct a classifier separating the set $\mathcal{X}$ from $\mathcal{Y}$.

According to theoretical constructions presented in the previous section, the following is an advantage for successful and efficient separation of random sets in high dimension: one of the sets (set $\mathcal{Y}$) should be sufficiently concentrated (spatially localized and have an exponentially smaller volume relative to the other [Theorems \ref{thm:one_separability}, \ref{thm:set_separability}]). If this is the case then, successful separability of this set of smaller volume depends on absence of unexpected concentrations in the probability distributions. Importantly, the probability of success approaches one exponentially fast, as a function of the data dimensionality.

In practice, however, the assumption that one of the sets is spatially localized in a small volume is too restrictive. To overcome this issue, we propose to partition/cluster the  set $\mathcal{Y}$ into a union of spatially localized subsets. Presence of local concentrations and separability issues have been linked and analyzed in \cite{GorMakTyu:2018}, \cite{gorban2018correction}, \cite{Zinovyev:IJCNN:2019}. The proposed clustering of the set $\mathcal{Y}$ aims at addressing these issues too.

Below we present an algorithm for  fast and efficient error correction of AI systems which is motivated by these observations and intuition stemming from our theoretical results.


\begin{algorithm}\label{alg:fast} ($1-nn$ removal of AI errors. Training). Input: sets $\mathcal{X}$, $\mathcal{Y}$, the number of clusters, $k$, threshold, $\theta$ (or thresholds $\theta_1,\dots,\theta_k$).

	\begin{enumerate}
		\item Determining the centroid  $\bar\bfx$ of the $\mathcal{X}$. Generate two sets, $\mathcal{X}_c$, the centralized set $\mathcal{X}$, and $\mathcal{Y}^{\ast}$, the set obtained from $\mathcal{Y}$ by subtracting $\bar\bfx$ from each of its elements.
		\item  Construct Principal Components for the centralized set $\mathcal{X}_c$.
		\item  Using Kaiser, broken stick, conditioning rule, or otherwise, select $m\leq n$ Principal Components, $h_1,\dots,h_m$, corresponding to the first largest eivenvalues $\lambda_1\geq \cdots \geq \lambda_m >0$ of the covariance matrix of the set $\mathcal{X}_c$, and project the centralized set $\mathcal{X}_c$ as well as $\mathcal{Y}^{\ast}$ onto these vectors. The operation returns sets $\mathcal{X}_r$ and $\mathcal{Y}_{r}^{\ast}$, respectively:
	\[
		\begin{aligned}
		\mathcal{X}_r&=\{\bfx | \bfx= H \bfz, \ \bfz\in\mathcal{X}_c\}\\
		\mathcal{Y}_r^\ast&=\{\bfy | \bfy=H \bfz, \ \bfz\in\mathcal{Y}^\ast\}, \ H=\left(\begin{array}{c} h_1^T \\ \vdots \\ h_m^T\end{array}\right).
		\end{aligned}
		\]
		\item Construct matrix $W$
	\[
	W=\mathrm{diag} \left(\frac{1}{\sqrt{\lambda_1}},\dots,\frac{1}{\sqrt{\lambda_m}}\right)
	\]
corresponding to the whitening transformation for the set $\mathcal{X}_r$. Apply the whitening transformation to sets $\mathcal{X}_r$ and $\mathcal{Y}_{r}^\ast$. This returns sets $\mathcal{X}_w$ and $\mathcal{Y}_{w}^{\ast}$:
	\[
	\begin{aligned}
		\mathcal{X}_w&=\{\bfx | \bfx= W \bfz, \ \bfz\in\mathcal{X}_r\}\\
		\mathcal{Y}_w^\ast&=\{\bfy | \bfy=W \bfz, \ \bfz\in\mathcal{Y}_r^\ast\}.
		\end{aligned}
	\]
		\item Cluster the set $\mathcal{Y}_{w}^\ast$ into $k$ clusters $\mathcal{Y}_{w,1}^\ast,\dots,\mathcal{Y}_{w,k}^\ast$ . Let $\bar{\bfy}_1,\dots,\bar{\bfy}_k$ be their corresponding centroids.
		\item  For each pair  $(\mathcal{X}_w,\mathcal{Y}_{w,i}^\ast)$, $i=1,\dots,k$, construct (normalized)  Fisher discriminants  $\bfw_1,\dots,\bfw_k$:
		\[
		\bfw_i=- \frac{(\Cov(\mathcal{X}_w)+\Cov(\mathcal{Y}_{w,i}^\ast))^{-1} \bar{\bfy}_i }{\|(\Cov(\mathcal{X}_w)+\Cov(\mathcal{Y}_{w,i}^\ast))^{-1} \bar{\bfy}_i \|}.
		\]
An element $\bfz$ is associated with the set $\mathcal{Y}_{w,i}^\ast$ if $(\bfw_i,\bfz)<\theta$ and with the set  $\mathcal{X}_w$ if $(\bfw_r,\bfz)\geq\theta$. 

If multiple thresholds are given then an element $\bfz$ is associated with the set $\mathcal{Y}_{w,i}^\ast$ if $(\bfw_i,\bfz)<\theta_i$ and with the set  $\mathcal{X}_w$ if $(\bfw_r,\bfz)\geq\theta_i$.

	\end{enumerate}
Output: vectors $\bfw_i$, $i=1,\dots,k$, matrices $H$ and $W$.
\end{algorithm}

The deployment/application part of the algorithm is as follows:

\begin{algorithm}\label{alg:fast:test} ($1-nn$ removal of AI errors. Deployment). Input: a data vector $\boldsymbol{x}$, the set's $\mathcal{X}$ centroid vector $\bar{\bfx}$, matrices $H$, $W$,  the number of clusters, $k$, cluster centroids $\bar{\bfy}_1,\dots,\bar{\bfy}_k$,  threshold, $\theta$ (or thresholds $\theta_1, \dots, \theta_k$), discriminant vectors, $\bfw_i$, $i=1,\dots,k$.

	\begin{enumerate}
		\item  Compute
		\[
		\bfx_w=W H (\bfx-\bar{\bfx})
		\]
		\item  Determine
		\[
		\ell=\arg \min_{i} \|\bfx_w-\bar{\bfy}_i\|.
		\]
		\item   Associate the vector  $\bfx$ with the set $\mathcal{Y}$ if $(\bfw_{\ell},\bfx_w)<\theta$ and with the set $\mathcal{X}$ otherwise. 
		
		If multiple thresholds are given then associate the vector  $\bfx$ with the set $\mathcal{Y}$ if $(\bfw_{\ell},\bfx_w)<\theta_\ell$ and with the set $\mathcal{X}$ otherwise.
	\end{enumerate}
Output: a label attributed to the vector $\bfx$.
\end{algorithm}

In contrast to previously proposed approaches using stochastic separation effects  \cite{tyukin2017knowledge}, Algorithm \ref{alg:fast:test} mitigates the presence of clusters whose centroids are close to the origin. If such clusters do occur and the  fraction of the set $\mathcal{X}$  located in their vicinity is not overwhelmingly large (which is ensured by Theorem \ref{thm:dichotomy} and Corollary \ref{cor:dichotomy}) then, at the stage of deployment, the corresponding correcting discriminants will be triggered by elements from $\mathcal{X}$ infrequently. 

\begin{figure}[h!]
\includegraphics[scale=.3]{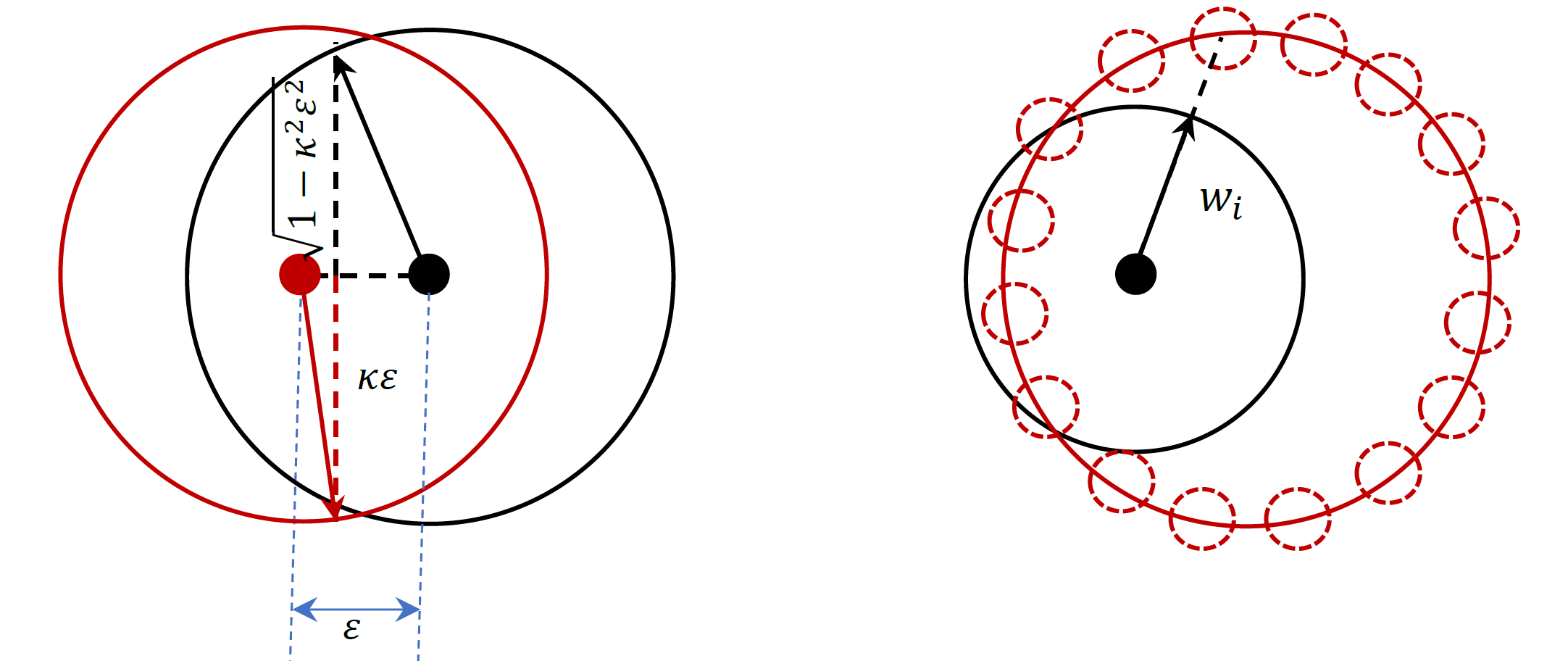}
\centering
\caption{Separation of a non-isolated set $\mathcal{Y}$ from $\mathcal{X}$. Black circle represents the set $\mathcal{X}$, red circle represents the set $\mathcal{Y}$. Filled disks represent centres of the sets $\mathcal{X}$ and $\mathcal{Y}$, respectively. {\it Left panel}. Single-cluster case. {\it Right panel.} Multiple-cluster case. According to Theorem \ref{thm:dichotomy}, the classes are ``hollow'' inside if Assumption 2 with appropriate parameters holds for their corresponding distributions.}
\label{fig:clusters}
\end{figure}

\begin{rem}\label{rem:epsilon-away} \normalfont According to Theorem \ref{thm:one_separability}, a single-point set $\mathcal{Y}$ under some mild hypotheses would be separated from $\mathcal{X}$ with probability close to one. 

If the set $\mathcal{Y}$ consists of multiple correlated subsets then, as the number of clusters increases, one would expect that the algorithm's performance in separating the sets $\mathcal{X}$, $\mathcal{Y}$ improves. 

At the same time, one may not necessarily require a near-perfect separability. For  example, removal of $90\%$ of all errors at the cost of a slight performance degradation of the AI's basic functionality may be an acceptable compromise in many applications. If the data dimensionality is sufficiently high then the desired separation might be achieved with just a single linear functional, provided that the centroid $\bar{\bfy}$ of the set $\mathcal{Y}$ is separated away from the centroid $\bar{\bfx}$ of the set $\mathcal{X}$.

The rationale behind this observation is as follows. Let $\mathcal{X}$ be equidistributed in $\mathbb{B}_n$ and $\mathcal{Y}$ be drawn from another equidistribution in a unit $n$-ball but centered at a point whose Euclidean norm is $0<\varepsilon\ll1$ (see Fig. \ref{fig:clusters}). Let $\bfw = (\bar{\bfx} - \bar{\bfy})/\|\bar{\bfx} - \bar{\bfy}\| = - \varepsilon^{-1} \bar{\bfy}$. Let $\kappa \in (0,1)$, and let  $h(\bfx)=(\bfx,\bfw) + \kappa \varepsilon$ be the separating hyperplane so that if $h(\bfz)>0$ then the vector $\bfz$ is associated with $\mathcal{X}$, and $\bfz$ is associated with the set $\mathcal{Y}$ if $h(\bfz)\leq 0$. Then the fraction of elements from $\mathcal{X}$ ``missed'' (false negative response) by this rule is bounded from above by
\[
\rho_x=\left(1-\kappa^2{\varepsilon}^2\right)^\frac{n}{2},
\]
and the fraction of elements from $\mathcal{Y}$ incorrectly attributed to $\mathcal{X}$ (false positive response) is bounded from above by 
\[
\rho_y=\left(1-(1-\kappa)^2{\varepsilon}^2\right)^\frac{n}{2}.
\]
Hence, when $n$ is sufficiently large, both $\rho_x$, $\rho_y$ may be made acceptably small even if $\varepsilon$ is small too (cf \cite{gorban2020high}).
\end{rem}

\begin{rem} \normalfont Note that if the clustering step in Algorithm \ref{alg:fast} is performed so that, for every $i$
\[
 \bfz\in\mathcal{Y}_{w,i}^\ast \Rightarrow \|\bfz-\bar\bfy_i\| < \|\bfz-\bar\bfy_j\|, \ \mbox{for all} \ j\neq i
\]
 then the proposed 1-nn integration logic, as in Algorithm \ref{alg:fast:test}, correctly assigns elements from $\mathcal{Y}$ to their corresponding discriminants $\bfw_i$. For each query point, only one of the linear discriminants is active. This is markedly different from the more aggressive ``union'' integration logic (OR rule) proposed in \cite{GorTyuGreen2019} in which, for a single query point, all discriminants are active simultaneously leading to higher chances of producing false negative errors.  In this respect, the 1-nn rule is somewhat tighter than the OR rule.
\end{rem}

\begin{rem} \normalfont It may sometimes be computationally advantageous to perform the clustering step in Algorithm \ref{alg:fast} (step 5) prior to dimensionality reduction. This will result in that the deployment part of the algorithm, Algorithm \ref{alg:fast:test} changes as follows
\begin{algorithm}\label{alg:fast:test:v2} ($1-nn$ removal of AI errors. Deployment) Input: a data vector $\boldsymbol{x}$, the set's $\mathcal{X}$ centroid vector $\bar{\bfx}$, matrices $H$, $W$,  the number of clusters, $k$, cluster centroids $\bar{\bfy}_1,\dots,\bar{\bfy}_k$,  threshold, $\theta$ (or thresholds $\theta_1,\dots,\theta_k$), discriminant vectors, $\bfw_i$, $i=1,\dots,k$.

Pre-compute vectors
\[
\bfw^\ast_{i}=H^T W  \bfw_i.
\]

\begin{enumerate}
		\item  Determine
		\[
		\ell=\arg \min_{i} \|\bfx - \bar{\bfx} - \bar{\bfy}_i\|.
		\]
		\item   Associate the vector  $\bfx$ with the set $\mathcal{Y}$ if $(\bfw_{\ell}^\ast,(\bfx-\bar{\bfx}))<\theta$ and with the set $\mathcal{X}$ otherwise.
		
		If multiple thresholds are given then associate the vector  $\bfx$ with the set $\mathcal{Y}$ if $(\bfw_{\ell}^\ast,(\bfx-\bar{\bfx}))<\theta_\ell$ and with the set $\mathcal{X}$ otherwise.
		
\end{enumerate}
\end{algorithm}

The difference between Algorithm \ref{alg:fast:test} and \ref{alg:fast:test:v2} is that the data point $\bfx$ no longer needs to be projected onto the principal components. This may be computationally advantageous when the number of components on which the data is projected is larger than the number of clusters $k$.
\end{rem}

Next section illustrates application of the algorithms with in two practical scenarios of edge-based AI deployment.

\section{Examples}\label{sec:examples}

The choice of examples is motivated primarily by our intention to illustrate the application of the new algorithm with and without clustering. The first example illustrates how one can take advantage of high-dimensional data for constructing a single discriminant. The second example enables us to show advantages of clustering for a problem where the feature space is genuinely high-dimensional and where the number of errors made by edge-based systems necessitates automated and computationally efficient interventions to be deployed at the edge.

\subsection{Real-time Tool Wear and Product Quality Prediction  for Computer Numerical Controlled (CNC) Milling Machines}

\subsubsection{System overview and setup}

Milling is a process of removing excess material by advancing a cutter into a work piece. It is one of the most commonly used processes for machining  freeform surfaces and custom parts \cite{lasemi2010recent}. Quality of the final part depends on on many factors including  tool path, tool orientation, tool geometry, tool wear and security of the part fixing. Here we focused on real-time prediction of part quality from the measurements characterizing electrical and mechanical state of the CNC machine.

\begin{figure}[h!]
\includegraphics[scale=.2]{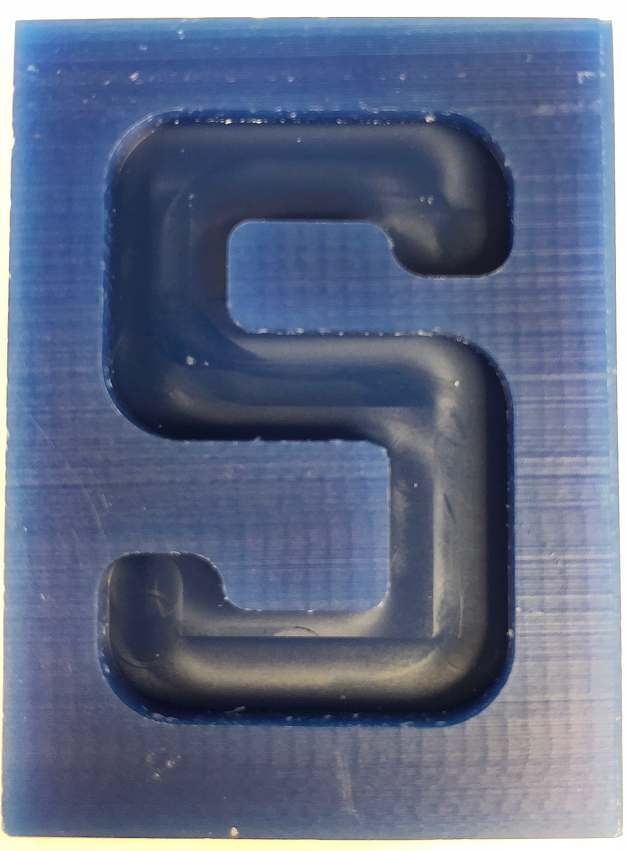}
\centering
\caption{An example of a part produced by the CNC milling machine \cite{tools}. }
\label{fig:part}
\end{figure}

To build a data-driven quality detector we used the CNC Milling Dataset from the University of Michigan Smart Lab  \cite{tools}. The dataset  contains a  series of machining experiments  run on 2" x 2" x 1.5" wax blocks. Machining data was collected from a CNC machine for variations of tool condition, feed rate, and clamping pressure. An example of a finished wax part with an "S" shape - S for smart manufacturing - carved into the top face is shown in Fig. \ref{fig:part}. 

Time series data from the machine was collected from $18$ experiments with a sampling rate of $100$ ms, and each instantaneous measurement vector contained $48$ attributes. The task was to predict the output quality of the part, as confirmed by visual inspection, from the instantaneous measurements.

\subsubsection{Automated Part Quality Prediction}

In this problem, we used Algorithms \ref{alg:fast}, \ref{alg:fast:test} with a single cluster. We also used a slightly modified Step 3 in  \ref{alg:fast} in which we retained $21$ Principal Components: from the $20$th to the $40$th. Note that we did not use the first $19$ components as in this particular problem inclusion of these components did translate into noticeable changes of the model's performance. 

{\it Training and validation datasets}. To train the model we used $9$ experiments (out of the total $18$) in which $5$ experiments corresponded to machining with unworn tool and which passed a visual inspection (experiments $1$, $3$, $4$,$5$,$11$) and $4$ experiments contained data corresponding to runs that either did not finish or where the part did not pass visual inspection (experiments $6$, $7$, $8$, $9$); $3$ experiments were used for testing (experiments $2$ and $17$ corresponding to successful runs where parts passed visual inspection, and experiment $10$ in which the part failed the inspection).

\begin{figure}[h!]
\includegraphics[width=0.7\textwidth]{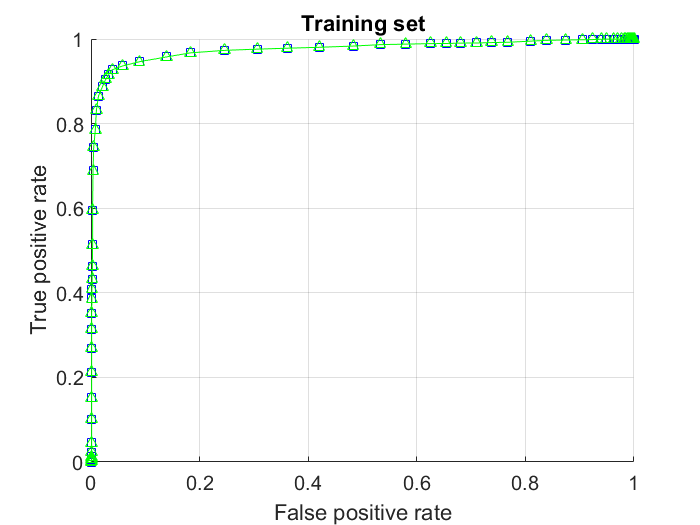}
\includegraphics[width=0.7\textwidth]{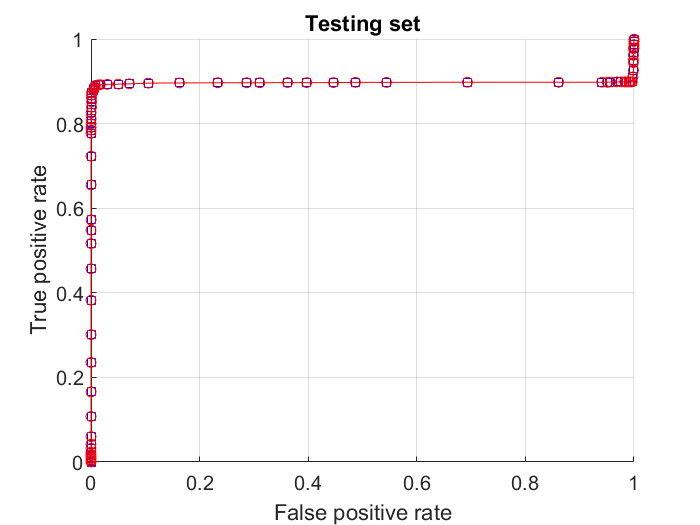}
\centering
\caption{ROC curves for detecting a faulty run from a single measurement in a run.}
\label{fig:training_testing_ROC_tool}
\end{figure}

{\it Experiments and results.} Each measurement in both training and testing datasets have been labeled as ``pass'' or ``fail'' depending on whether the run from which this measurement was taken passed the visual inspection (and hence the point was labeled as a ``pass'') or failed (resulting in the label ``fail''). Training of the model took $0.16$ to $0.2$ seconds on a core i7 laptop, and a summary of the model performance is summarized in Fig. \ref{fig:training_testing_ROC_tool}.

\begin{figure}[h!]
\includegraphics[width=0.7\textwidth]{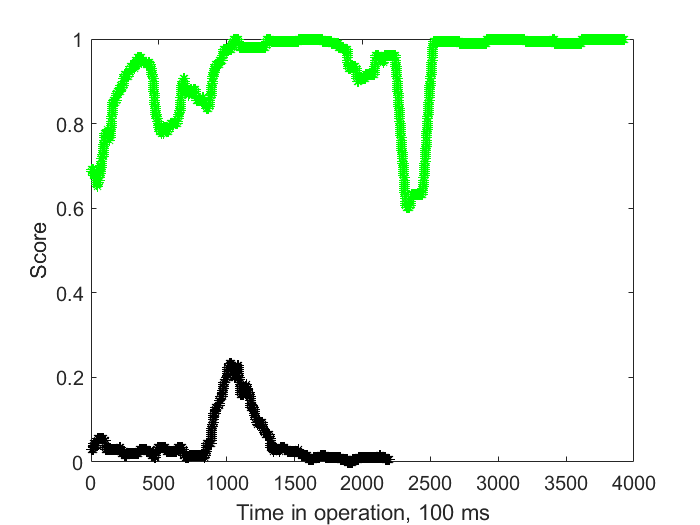}
\includegraphics[width=0.7\textwidth]{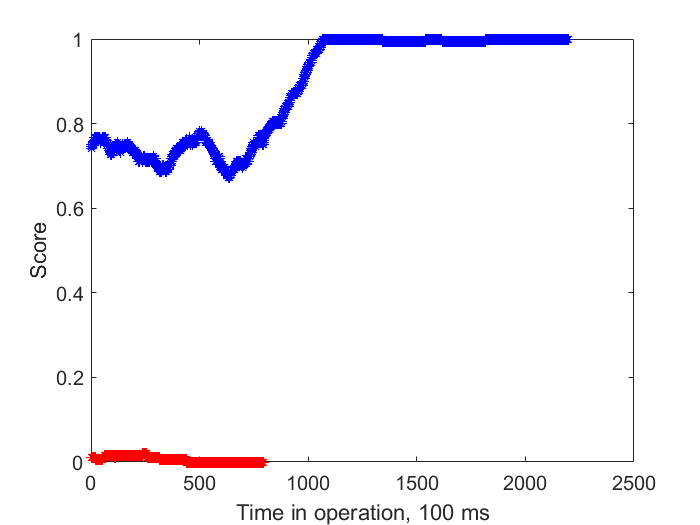}
\centering
\caption{Averaged prediction ``Score'' for training (top panel) and testing runs (bottom panel). Green and black curves in the top panel show correct and failed machining for the training set, respectively. Blue and red curves correspond to correct and failed runs in the test set, respectively.} \label{fig:training_testing_tool}
\end{figure}

To better integrate the model's decision and filter spurious errors we averaged the model's binary output (computed with the threshold of $-1.5$) over a sliding window of $200$ measurements. The resulting value for each window was recorded and shown as  ``Score'' in Fig. \ref{fig:training_testing_tool}.  Outcomes of these experiments on the entire training and test sets,  where the data from several runs was simply concatenated, are presented in Fig. \ref{fig:training_testing_tool}.

As one can see from these figures, the model correctly identifies failed runs. Good generalization performance of this simple model can be explained by the concentration effects captured in Fig. \ref{fig:clusters}: relatively small differences of class means are apparently sufficient to ensure reasonable class separation by a Fisher discriminant if the data dimension is sufficiently large. 

\begin{figure}[h!]
\includegraphics[width=0.7\textwidth]{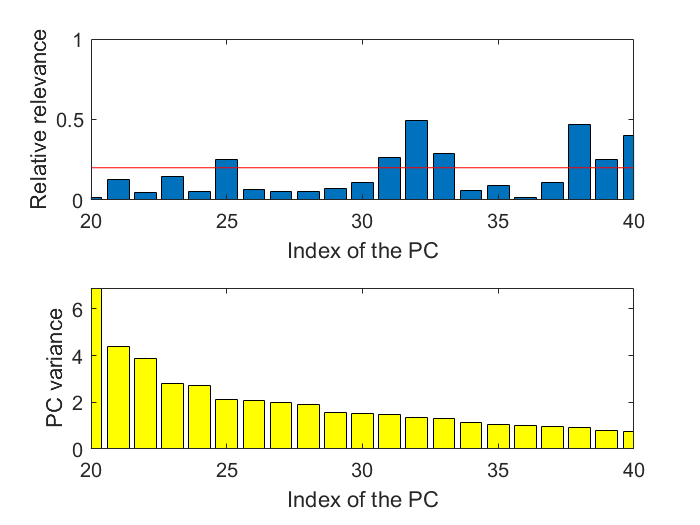}
\centering
\caption{Relative relevance of Principal Components for class discrimination.} \label{fig:relevant_relevance}
\end{figure}

In addition to this, we notw that not every single dimension is equally important either. To illustrate this point, let $\bar{\bfx}$ and $\bar{\bfy}$ be empirical class means for the ``pass'' and ``fail'' classes. We calculated 
\[
\mbox{Relative relevance}_i=\left|\frac{\bar{\bfx}-\bar{\bfy}}{\|\bar{\bfx}-\bar{\bfy}\|} \frac{h_i}{\|h_i|}\right|, \ i=20,\dots,40,
\]
where $h_i$ are the corresponding Principal Components (Algorithm \ref{alg:fast}). The relative relevance indices  for each $i$-th component and the contribution of that component to the total empirical data variance are shown in Fig. \ref{fig:relevant_relevance}.
As we can see from Fig. \ref{fig:relevant_relevance}, a large proportion of principal components only marginally project onto the vector $\bar{\bfx}-\bar{\bfy}$ which further contributes to separability.

\subsection{Adaptive Removal of False Positives in Video Surveillance and Analytics Systems: A Case Study}
\subsubsection{System overview and setup}

All UK airports are expected to ensure that the average passenger spends no longer than twelve minutes going through the security area.

The current boarding gates can measure the number of passengers coming through the security area via passenger boarding passes. However there is no way of determining whether an individual passenger has left the security area.
The current solution involves a member of staff manually keeping track of a small sample of passengers passing from the boarding gates to the security scanners and logging the time taken.

Knowledge about the length of these queues, as well as the number of passengers getting through the airport,  helps airports to manage their resources in an efficient way by enabling them to decide how many security stations should be open. It also provides passengers with valuable information on the amount of time  they can expect to spend inside the queues thereby allowing them to manage their time inside the airport more efficiently. Thus, knowing the time taken for a passenger since entering boarding-pass gates until leaving security gates, in almost real time, would be very beneficial. However, the current practices aimed at addressing this specific problem are far from being efficient.

To address the efficiency issue, various Queue Management System (QMS) are being developed. Here we provide a short description of a system which has been developed by the Visual Management Systems Ltd within the scope of the Innovate UK Knowledge Transfer Partnership project (KTP 10522). 

The system consists of two major components: the hardware that is in charge of detecting faces, and the back-end server that processes the data streamed from the hardware unit and calculates the average, fastest and slowest security queue times. A front-end web-page is served by the back-end to display the aforementioned statistical data as well as producing historical reports. As mentioned before, security queue time is the time a passenger spends in queues for the Boarding Pass Gates (BPG) and Security Gates (SG). 

\begin{figure}[h!]
\includegraphics[scale=.4]{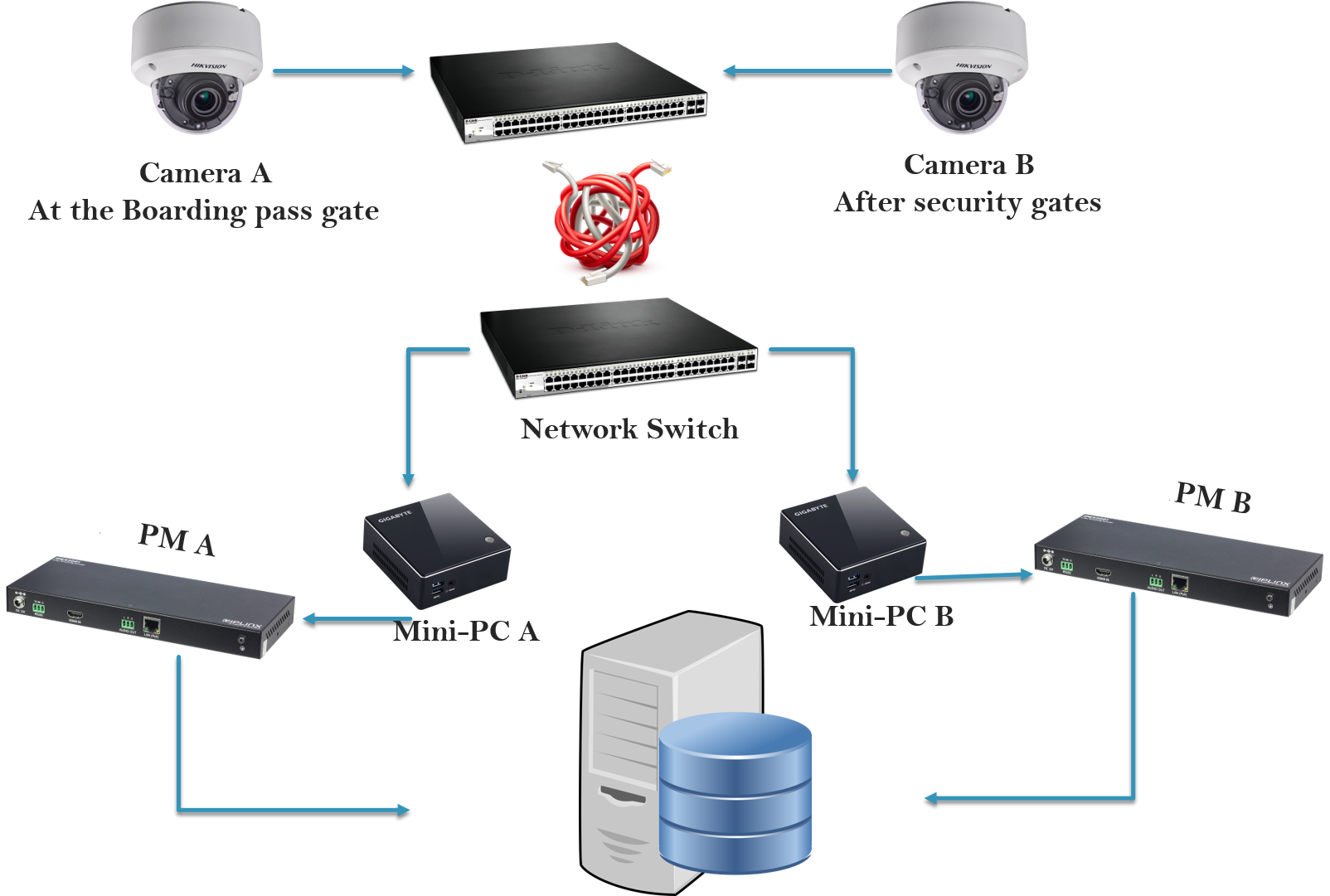}
\centering
\caption{Queue Management System}
\label{fig:hardware}
\end{figure}

For the hardware, shown in Figure \ref{fig:hardware}, we utilise two (or more) high-definiton cameras streaming in H.264, one for BPG and one for SG, connected to two \emph{Processing Modules \textsuperscript{TM}} (PM)  via two mini PC's.

The function of the PMs is to detect and track faces from the video feeds and output relevant metadata (face coordinates, time stamp, bitmap/image of the face etc) to the processing server. The metadata are then received by the back-end server for processing. The  back-end includes four different sections; parsing, image pre-processing, feature extraction and the data base. In the parsing section, the encoded face bimap is extractedl together with all other metadata. This is followed by the pre-processing unit in which each thumbnail is encoded to JPG format and saved to the local hard disk before passing through the Dlib library to generate a face score. Then, the thumbnail will pass through a Convolutional Neural Network, e.g. VGG-16 \cite{Simonyan:2015}, for feature extraction. Finally, features of the thumbnail together with all related metadata will be stored in the database.

The system has been trialled over a period of $3$ months in a major UK airport. The trials revealed that, depending on operational conditions, PMs based on propriertrary algorithms, occasionally return false positive detects These false positive detects if left untreated, have a capacity to slow down the entire processing pipeline. As the system scales up, dealing with these false posiive detects on the side of the server becomes computationally prohibitive. Moreover, the false positive detects, are camera and place-specific, in general. Thus there is a need and a rationale to address these false positive detects at their source.

\subsubsection{Adaptive Removal of False Positives}

In order to address the problem of false positives, we implemented and tested the proposed algorithms in this setup. For the purposes of avoiding issues with reproducibility as well as due to the data protection, in what follows we present a detailed account of this implementation in which the PM  was an OpenCV implementation of the Haar face detector. The detector has been applied to a publically accessible video footage capturing traffic and pedestrians walking on the streets of Montreal. For the purposes of testing and validation, we used the MTCNN face detector as a vehicle to generate ground truth data. All the data as well as the code generating true positive and false positive images can be provided by request.

\begin{figure}
\centering
\includegraphics[width=0.7\textwidth]{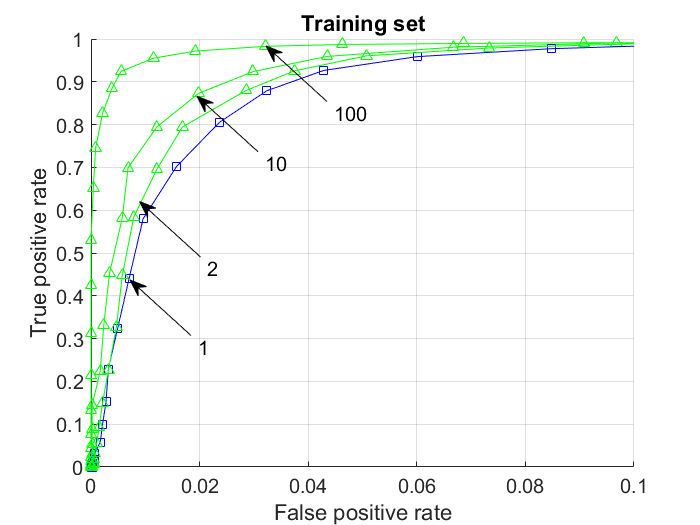}
\includegraphics[width=0.7\textwidth]{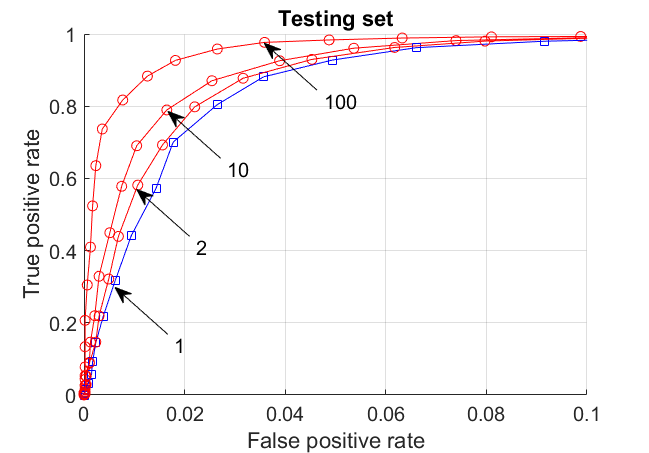}
\caption{ROC curves after the application of the AI error correcting algorithm with 200 Pricipal Components}\label{fig:1}
\end{figure}

For this particular dataset, the total number of true positives was $21896$, and the total number of false positives was $9372$. All the detects have been resized to $64\times 64$ crops (in RGB encoding). Each crop produces a $12288$-dimensional vector. From this dataset, we generated a training set containing $50$ percent of positive and false positives, and passed this training set to Algorithm  \ref{alg:fast}. In the algorithm, true positives have been associated with the set $\mathcal{X}$, and false positives were associated with the set $\mathcal{Y}_{\ast}$. The number of Principal Components was limited to $200$. We tried the algorithm for the following numbers of clusters:   $1$, $5$,$10$, and $100$ . At the deployment stage, we used Algorithm \ref{alg:fast:test}.  Training took, on average, about $180$ seconds on a Core i7 laptop, and the outcomes of the process as well as performance on the testing set are summarized in Fig. \ref{fig:1} where curves corresponding to different numbers of clusters ($1$, $5$,$10$, and $100$) are annotated by arrows with numbers $1$, $5$,$10$, and $100$, respectively.

As we can see from this figure, even a single-cluster  implementation of Algorithm \ref{alg:fast} allows one to filter 90 percent of all errors at the cost of missing circa 5 percent of true positives. This is consistent with expectations discussed in Remark \ref{rem:epsilon-away}.  Implementation of the single-cluster correcting functional on an ARM Cortex-A53 processor took less than 1 millisecond per  each 12288-dimensional vector implying significant capacity of the approach for embedded near-edge applications. 

A notable classification performance gain is observed for a $100$-cluster version of the algorithm. This, however, comes at additional computational costs at the stage of deployment. Having said this, the deployment part of the algorithm is extremely scalable leading to significant expected reductions of computation times in the case of parallel execution of the code and is hence amenable to massively parallel implementations.

\begin{figure}
\centering
\includegraphics[width=0.7\textwidth]{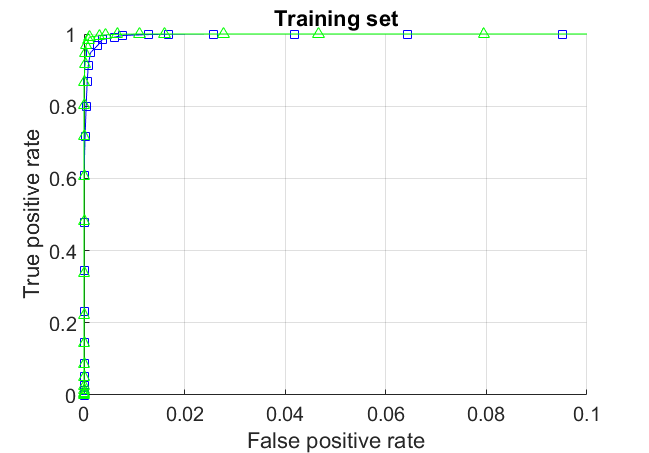}
\includegraphics[width=0.7\textwidth]{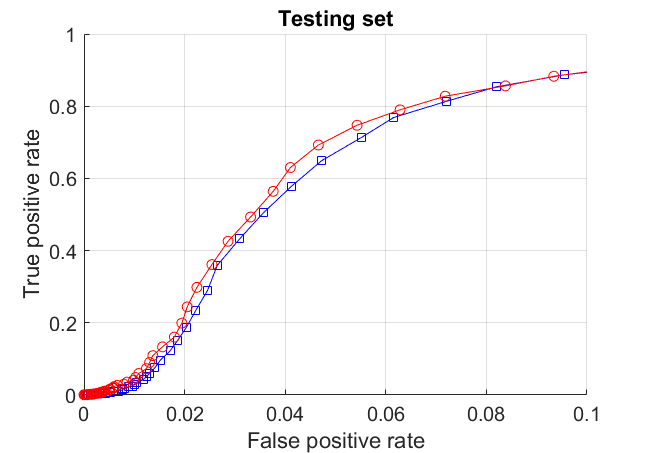}
\caption{ROC curves after the application of the AI error correcting algorithm with 6000 Principal Components. Blue curves correspond to the baseline case with a single cluster. Red and green curves show behavior of the system for $100$ clusters.}\label{fig:2}
\end{figure}

It is also worthwhile to mention that the concentration effects, as formulated in Theorems \ref{thm:one_separability}, \ref{thm:set_separability}, and which are at the backbone of Algorithms \ref{alg:fast} -- \ref{alg:fast:test:v2}, may negatively effect the overall system's performance if the dimensionality is excessively high and the cardinality of the set $\mathcal{Y}$ is comparable to that of the set $\mathcal{X}$.  To illustrate this point, we used Algorithms \ref{alg:fast}, \ref{alg:fast:test} with the $6000$ Principal Components. Results are shown in Fig. \ref{fig:2}

As we can see from this figure, if the retained dimensionality of the decision-making space is too large, the algorithms tend to overfit and hence special consideration needs to be given to the choice of the numbers of projections used and the volume of training data.

\section{Conclusion}
\label{sec:conclusion}

In this work we presented a novel approach for equipping edge-based or near-edge devices with capabilities to quickly learn on-the-job and continuously improve over time in presence of spurious as well as a rather overwhelming number of errors.  The approach is based on stochastic separation theorems \cite{GorbanTyukin:NN:2017}, \cite{GorTyuRom2016}, \cite{GorTyuRom2019}, \cite{tyukin2017knowledge}, \cite{gorban2018correction}, \cite{GorTyuGreen2019} and the concentration of measure phenomena. 

Our results demonstrate that the new capability can be delivered to the edge and deployed in a fully automated way, whereby a more sophisticated AI system monitors performance of a less powerful counterpart. The approach, for the first time uses $1-nn$ integration rule for error correction, as opposed to mere disjunctions. The $1-nn$ rule is justified by the dichotomy of high-dimensional datasets captured in Theorem \ref{thm:dichotomy} and Corollary \ref{cor:dichotomy}.  Experimentally, we investigated the sensitivity of the algorithm to change of its meta-parameters like the number of clusters and projections used. The results directly respond to the fundamental challenge of removing AI errors in industrial applications at minimal computational costs, and  some elements of the theory already underpin two US patents \cite{patent1}, \cite{patent2}.

Theoretical results are illustrated with two industrial applications: performance monitoring of the CNC milling processes and edge-based object detection.  An application field of the approach could be the class of randomized computational architectures such as stochastic configuration networks \cite{wang2017stochastic}, and in particular the employment of the measure concentration effects for estimating their approximation convergence rates. This will be the subject of our future work.

\section*{Acknowledgment}
\label{sec:ack}

The work was supported by Innovate UK Knowledge Transfer Partnership grant KTP010522, by the grant of the Russian Science Foundation (Project No. 19-19-00566), and by a 111 Project (B16009).



\bibliographystyle{abbrvnat}
\bibliography{fast_corrector_at_the_edge}

\begin{thebibliography}{38}
\providecommand{\natexlab}[1]{#1}
\providecommand{\url}[1]{\texttt{#1}}
\expandafter\ifx\csname urlstyle\endcsname\relax
  \providecommand{\doi}[1]{doi: #1}\else
  \providecommand{\doi}{doi: \begingroup \urlstyle{rm}\Url}\fi

\bibitem[Albergante et~al.(2019)Albergante, Bac, and
  Zinovyev]{Zinovyev:IJCNN:2019}
A.~Albergante, J.~Bac, and A.~Zinovyev.
\newblock Estimating the effective dimension of large biological datasets using
  fisher separability analysis.
\newblock In \emph{Proc. of the International Joint Conference on Neural
  Networks (IJCNN)}, 2019.

\bibitem[Chapelle(2007)]{chapelle2007training}
O.~Chapelle.
\newblock Training a support vector machine in the primal.
\newblock \emph{Neural computation}, 19\penalty0 (5):\penalty0 1155--1178,
  2007.

\bibitem[Chen et~al.(2015)Chen, Goodfellow, and Shlens]{chen2015net2net}
T.~Chen, I.~Goodfellow, and J.~Shlens.
\newblock Net2net: Accelerating learning via knowledge transfer.
\newblock \emph{International Conference on Learning Representations (ICLR)
  2016}, 2015.

\bibitem[Gorban and Tyukin(2017)]{GorbanTyukin:NN:2017}
A.~Gorban and I.~Tyukin.
\newblock Stochastic separation theorems.
\newblock \emph{Neural Networks}, 94:\penalty0 255--259, 2017.

\bibitem[Gorban and Tyukin(2018)]{gorban2018blessing}
A.~Gorban and I.~Tyukin.
\newblock Blessing of dimensionality: mathematical foundations of the
  statistical physics of data.
\newblock \emph{Philosophical Transactions of the Royal Society A},
  376:\penalty0 20170237, 2018.
\newblock \doi{10.1098/rsta.2017.0237}.

\bibitem[Gorban et~al.(2016{\natexlab{a}})Gorban, Tyukin, Prokhorov, and
  Sofeikov]{GorTyu:2016}
A.~Gorban, I.~Tyukin, D.~Prokhorov, and K.~Sofeikov.
\newblock Approximation with random bases: Pro et contra.
\newblock \emph{Information Sciences}, 364--365:\penalty0 129--145,
  2016{\natexlab{a}}.

\bibitem[Gorban et~al.(2016{\natexlab{b}})Gorban, Tyukin, and
  Romanenko]{GorTyuRom2016}
A.~Gorban, I.~Tyukin, and I.~Romanenko.
\newblock The blessing of dimensionality: Separation theorems in the
  thermodynamic limit, 09 2016{\natexlab{b}}.
\newblock URL \url{https://arxiv.org/abs/1610.00494v1}.
\newblock 2nd IFAC Workshop on Thermodynamic Foundations of Mathematical
  Systems Theory. September 28-30, 2016, Vigo, Spain.

\bibitem[Gorban et~al.(2018{\natexlab{a}})Gorban, Golubkov, Grechuk, Mirkes,
  and Tyukin]{gorban2018correction}
A.~Gorban, A.~Golubkov, B.~Grechuk, E.~Mirkes, and I.~Tyukin.
\newblock Correction of ai systems by linear discriminants: Probabilistic
  foundations.
\newblock \emph{Information Sciences}, 466:\penalty0 303--322,
  2018{\natexlab{a}}.

\bibitem[Gorban et~al.(2019)Gorban, Burton, Romanenko, and
  Tyukin]{GorTyuRom2019}
A.~Gorban, R.~Burton, I.~Romanenko, and I.~Tyukin.
\newblock One-trial correction of legacy {AI} systems and stochastic separation
  theorems.
\newblock \emph{Information Sciences}, 484:\penalty0 237--254, 2019.

\bibitem[Gorban et~al.(2018{\natexlab{b}})Gorban, Grechuk, and
  Tyukin]{gorban2018augmented}
A.~N. Gorban, B.~Grechuk, and I.~Y. Tyukin.
\newblock Augmented artificial intelligence.
\newblock \emph{arXiv preprint arXiv:1802.02172}, 2018{\natexlab{b}}.

\bibitem[Gorban et~al.(2018{\natexlab{c}})Gorban, Makarov, and
  Tyukin]{GorMakTyu:2018}
A.~N. Gorban, V.~A. Makarov, and I.~Y. Tyukin.
\newblock The unreasonable effectiveness of small neural ensembles in
  high-dimensional brain.
\newblock \emph{Physics of Life Reviews}, 2018{\natexlab{c}}.
\newblock \doi{10.1016/j.plrev.2018.09.005}.

\bibitem[Gorban et~al.(2020)Gorban, Makarov, and Tyukin]{gorban2020high}
A.~N. Gorban, V.~Makarov, and I.~Tyukin.
\newblock High-dimensional brain in a high-dimensional world: Blessing of
  dimensionality.
\newblock \emph{Entropy}, 22\penalty0 (1):\penalty0 82, 2020.

\bibitem[Gromov(2003)]{GAFA:Gromov:2003}
M.~Gromov.
\newblock Isoperimetry of waists and concentration of maps.
\newblock \emph{GAFA, Geomteric and Functional Analysis}, 13:\penalty0
  178--215, 2003.

\bibitem[{Hains} et~al.(2018){Hains}, {Jakobsson}, and {Khmelevsky}]{8369576}
G.~{Hains}, A.~{Jakobsson}, and Y.~{Khmelevsky}.
\newblock Towards formal methods and software engineering for deep learning:
  Security, safety and productivity for dl systems development.
\newblock In \emph{2018 Annual IEEE International Systems Conference (SysCon)},
  pages 1--5, April 2018.
\newblock \doi{10.1109/SYSCON.2018.8369576}.

\bibitem[Hansen and Salamon(1990)]{Hansen:1990}
L.~K. Hansen and P.~Salamon.
\newblock Neural network ensembles.
\newblock \emph{{IEEE} {T}ransactions on Pattern Analyis and Machince
  Intelligence}, 12\penalty0 (10):\penalty0 993--1001, 1990.

\bibitem[Ho(1995)]{Ho:1995}
T.~K. Ho.
\newblock Random decision forests.
\newblock In \emph{Proc. of the 3rd International Conference on Document
  Analysis and Recognition}, pages 993--1001, 1995.

\bibitem[Ho(1998)]{Ho:1998}
T.~K. Ho.
\newblock The random subspace method for constructing decision forests.
\newblock \emph{{IEEE} {T}ransactions on Pattern Analyis and Machince
  Intelligence}, 20\penalty0 (8):\penalty0 832--844, 1998.

\bibitem[Kainen and Kurkova(1993)]{Kurkova}
P.~Kainen and V.~Kurkova.
\newblock Quasiorthogonal dimension of euclidian spaces.
\newblock \emph{Appl. Math. Lett.}, 6\penalty0 (3):\penalty0 7--10, 1993.

\bibitem[Kainen(1997)]{kainen1997utilizing}
P.~C. Kainen.
\newblock Utilizing geometric anomalies of high dimension: When complexity
  makes computation easier.
\newblock In \emph{Computer Intensive Methods in Control and Signal
  Processing}, pages 283--294. Springer, 1997.

\bibitem[Kuznetsova et~al.(2015)Kuznetsova, Hwang, Rosenhahn, and
  Sigal]{Kuznetsova:2015}
A.~Kuznetsova, S.~Hwang, B.~Rosenhahn, and L.~Sigal.
\newblock Expanding object detector’s horizon: Incremental learning framework
  for object detection in videos.
\newblock In \emph{Proc. of the IEEE Conference on Computer Vision and Pattern
  Recognition (CVPR)}, pages 28--36, 2015.

\bibitem[Lasemi et~al.(2010)Lasemi, Xue, and Gu]{lasemi2010recent}
A.~Lasemi, D.~Xue, and P.~Gu.
\newblock Recent development in cnc machining of freeform surfaces: A
  state-of-the-art review.
\newblock \emph{Computer-Aided Design}, 42\penalty0 (7):\penalty0 641--654,
  2010.

\bibitem[{Li} et~al.(2018){Li}, {Oyler}, {Zhang}, {Yildiz}, {Kolmanovsky}, and
  {Girard}]{7993050}
N.~{Li}, D.~W. {Oyler}, M.~{Zhang}, Y.~{Yildiz}, I.~{Kolmanovsky}, and A.~R.
  {Girard}.
\newblock Game theoretic modeling of driver and vehicle interactions for
  verification and validation of autonomous vehicle control systems.
\newblock \emph{IEEE Transactions on Control Systems Technology}, 26\penalty0
  (5):\penalty0 1782--1797, Sep. 2018.
\newblock \doi{10.1109/TCST.2017.2723574}.

\bibitem[Liang et~al.(2019)Liang, Tsui, Ni, Valentim, Baxter, Liu, Cai,
  Kermany, Sun, Chen, He, Zhu, Tian, Shao, Zheng, Hou, Hewett, Li, Liang, Zang,
  Zhang, Pan, Cai, Ling, Li, Cui, Tang, Ye, Huang, He, Liang, Zhang, Jiang, Yu,
  Gao, Ou, Deng, Hou, Wang, Yao, Liang, Zhang, Duan, Zhang, Gibson, Zhang, Li,
  Zhang, Karin, Nguyen, Wu, Wen, Xu, Xu, Wang, Wang, Li, Pizzato, Bao, Xiang,
  He, He, Zhou, Haw, Goldbaum, Tremoulet, Hsu, Carter, Zhu, Zhang, and
  Xia]{Liang:2019}
H.~Liang, B.~Tsui, H.~Ni, C.~Valentim, S.~Baxter, G.~Liu, W.~Cai, D.~Kermany,
  K.~Sun, J.~Chen, L.~He, J.~Zhu, P.~Tian, H.~Shao, L.~Zheng, R.~Hou,
  S.~Hewett, G.~Li, P.~Liang, X.~Zang, Z.~Zhang, L.~Pan, H.~Cai, R.~Ling,
  S.~Li, Y.~Cui, S.~Tang, H.~Ye, X.~Huang, W.~He, W.~Liang, Q.~Zhang, J.~Jiang,
  W.~Yu, J.~Gao, W.~Ou, Y.~Deng, Q.~Hou, B.~Wang, C.~Yao, Y.~Liang, S.~Zhang,
  Y.~Duan, R.~Zhang, S.~Gibson, C.~Zhang, O.~Li, E.~Zhang, G.~Karin, N.~Nguyen,
  X.~Wu, C.~Wen, J.~Xu, W.~Xu, B.~Wang, W.~Wang, J.~Li, B.~Pizzato, C.~Bao,
  D.~Xiang, W.~He, S.~He, Y.~Zhou, W.~Haw, M.~Goldbaum, A.~Tremoulet, C.-N.
  Hsu, H.~Carter, L.~Zhu, K.~Zhang, and H.~Xia.
\newblock Evaluation and accurate diagnoses of pediatric diseases using
  artificial intelligence.
\newblock \emph{Nature Medicine}, 25:\penalty0 433--438, 2019.
\newblock \doi{https://doi.org/10.1038/s41591-018-0335-9}.

\bibitem[{Meltz} and {Guterman}(2019)]{8723533}
D.~{Meltz} and H.~{Guterman}.
\newblock Functional safety verification for autonomous ugvs—methodology
  presentation and implementation on a full-scale system.
\newblock \emph{IEEE Transactions on Intelligent Vehicles}, 4\penalty0
  (3):\penalty0 472--485, Sep. 2019.
\newblock \doi{10.1109/TIV.2019.2919460}.

\bibitem[Misra et~al.(2015)Misra, Shrivastava, and Hebert]{Misra:2015}
I.~Misra, A.~Shrivastava, and M.~Hebert.
\newblock Semi-supervised learning for object detectors from video.
\newblock In \emph{Proc. of the IEEE Conference on Computer Vision and Pattern
  Recognition (CVPR)}, pages 3594--3602, 2015.

\bibitem[Pratt(1992)]{Pratt:ANIP:1992}
L.~Pratt.
\newblock Discriminability-based transfer between neural networks.
\newblock \emph{Advances in Neural Information Processing Systems}, \penalty0
  (5):\penalty0 204--211, 1992.

\bibitem[Prest et~al.(2012)Prest, Leistner, Civera, Schmid, and
  Ferrari]{Prest:2012}
A.~Prest, C.~Leistner, J.~Civera, C.~Schmid, and V.~Ferrari.
\newblock Learning object class detectors from weakly annotated video.
\newblock In \emph{Proc. of the IEEE Conference on Computer Vision and Pattern
  Recognition (CVPR)}, pages 3282--3289, 2012.

\bibitem[Romanenko et~al.(2018)Romanenko, Tyukin, Gorban, and
  Sofeikov]{patent2}
I.~Romanenko, I.~Tyukin, A.~Gorban, and K.~Sofeikov.
\newblock Method of image processing. {US} patent {US}10062013{B}2, August, 28
  2018.
\newblock URL \url{https://patents.google.com/patent/US10062013B2/en}.

\bibitem[Romanenko et~al.(2019)Romanenko, Gorban, and Tyukin]{patent1}
I.~Romanenko, A.~Gorban, and I.~Tyukin.
\newblock Image processing. {US} patent {US}10489634{B}2, November, 26 2019.
\newblock URL \url{https://patents.google.com/patent/US20180089497A1/en}.

\bibitem[Simonyan and Zisserman(2015)]{Simonyan:2015}
K.~Simonyan and A.~Zisserman.
\newblock Very deep convolutional networks for large-scale image recognition.
\newblock In \emph{International Conference on Learning Representations}, 2015.
\newblock arXiv:1409.1556.

\bibitem[Sun(2018)]{tools}
S.~Sun.
\newblock {CNC} mill tool wear dataset, 2018.
\newblock URL
  \url{https://www.kaggle.com/shasun/tool-wear-detection-in-cnc-mill}.

\bibitem[{Takács} et~al.(2018){Takács}, {Rudas}, {Bösl}, and
  {Haidegger}]{8574003}
A.~{Takács}, I.~{Rudas}, D.~{Bösl}, and T.~{Haidegger}.
\newblock Highly automated vehicles and self-driving cars [industry tutorial].
\newblock \emph{IEEE Robotics Automation Magazine}, 25\penalty0 (4):\penalty0
  106--112, Dec 2018.
\newblock \doi{10.1109/MRA.2018.2874301}.

\bibitem[Tyukin et~al.(2019)Tyukin, Gorban, Green, and
  Prokhorov]{GorTyuGreen2019}
I.~Tyukin, A.~Gorban, S.~Green, and D.~Prokhorov.
\newblock Fast construction of correcting ensembles for legacy artificial
  intelligence systems: Algorithms and a case study.
\newblock \emph{Information Sciences}, 485:\penalty0 230--247, 2019.

\bibitem[Tyukin et~al.(2018)Tyukin, Gorban, Sofeikov, and
  Romanenko]{tyukin2017knowledge}
I.~Y. Tyukin, A.~N. Gorban, K.~Sofeikov, and I.~Romanenko.
\newblock Knowledge transfer between artificial intelligence systems.
\newblock \emph{Frontiers of Neurorobotics}, 12, Article 49, 2018.
\newblock \doi{10.3389/fnbot.2018.00049}.

\bibitem[Vapnik(2000)]{Vapnik2000}
V.~Vapnik.
\newblock \emph{The Nature of Statistical Learning Theory}.
\newblock Springer-Verlag, 2000.

\bibitem[Vapnik and Izmailov(2017)]{vapnik2017knowledge}
V.~Vapnik and R.~Izmailov.
\newblock Knowledge transfer in svm and neural networks.
\newblock \emph{Annals of Mathematics and Artificial Intelligence}, pages
  1--17, 2017.

\bibitem[Wang and Li(2017)]{wang2017stochastic}
D.~Wang and M.~Li.
\newblock Stochastic configuration networks: Fundamentals and algorithms.
\newblock \emph{IEEE Transactions on Cybernetics}, 47\penalty0 (10):\penalty0
  3466--3479, 2017.

\bibitem[Zheng et~al.(2016)Zheng, Song, Leung, and Goodfellow]{Zheng:2016}
S.~Zheng, Y.~Song, T.~Leung, and I.~Goodfellow.
\newblock Improving the robustness of deep neural networks via stability
  training.
\newblock In \emph{Proc. of the IEEE Conference on Computer Vision and Pattern
  Recognition (CVPR)}, 2016.
\newblock https://arxiv.org/abs/1604.04326.

\end{thebibliography}

\end{document}